\documentclass[letterpaper, 10 pt, conference]{ieeeconf}  

\IEEEoverridecommandlockouts                              

\usepackage{amsmath} 
\usepackage{hyperref}
\usepackage[font=footnotesize,labelfont=bf]{subcaption}
\usepackage{algorithm,algpseudocode}
\usepackage{cite}
\usepackage[pdftex]{graphicx}
\graphicspath{{../pdf/}{../jpeg/}}
\DeclareGraphicsExtensions{.pdf,.jpeg,.png}

\usepackage[para]{footmisc}

\makeatletter
\newcommand{\printfnsymbol}[1]{%
  \textsuperscript{\@fnsymbol{#1}}%
}
\title{\LARGE \bf
Robot Person Following in Uniform Crowd Environment
}

\author{Adarsh Ghimire\printfnsymbol{1}\thanks{*Authors make equal contributions to this work.},
Xiaoxiong Zhang\printfnsymbol{1},
Sajid Javed,
Jorge Dias,
Naoufel Werghi
\thanks{This work acknowledges the support provided by the Khalifa University of Science and Technology under award No. RC1-2018-KUCARS.}
}

\begin{document}

\maketitle
\thispagestyle{empty}
\pagestyle{empty}

\begin{abstract}
Person-tracking robots have many applications, such as in security, elderly care, and socializing robots. Such a task is particularly challenging when the person is moving in a Uniform crowd. Also, despite significant progress of trackers reported in the literature, state-of-the-art trackers have hardly addressed person following in such scenarios. 
In this work, we focus on improving the perceptivity of a robot for a person following task by developing a robust and real-time applicable object tracker. We present a new robot person tracking system with a new RGB-D tracker, \textit{D}eep \textit{T}racking with \textit{R}GB-\textit{D} (DTRD) that is resilient to tricky challenges introduced by the uniform crowd environment. Our tracker utilizes transformer encoder-decoder architecture with RGB and depth information to discriminate the target person from similar distractors. A substantial amount of comprehensive experiments and results demonstrate that our tracker has higher performance in two quantitative evaluation metrics and confirms its superiority over other SOTA trackers.
\end{abstract}

\section{INTRODUCTION}

In the recent decade, there has been a growing demand for mobile robots for a person following in the human-robot collaboration works such as assisting robots in industry, healthcare, surveillance, elderly care, and monitoring \cite{islam2019person}. Conventional robot tracking systems used lasers, which provide a cluster of distance measurements, to perceive the environment as well as to detect and track the target person \cite{islam2019person}. However, laser data are usually noisy and difficult to use for re-identification of the target. With the advent of deep learning, there has been a rapid growth of visual tracking algorithms \cite{VOT_TPAMI, kristan2018sixth, kristan2019seventh, kristan2020eighth, Kristan2021ninth}, which makes robot person following using vision-based trackers one of the popular and intelligent approaches \cite{chen2017integrating, chen2017person}.
The primary challenge is to accurately localize the target person of interest and take necessary control action to follow. Inherently, the task is quite challenging because it has to continuously recognize the target person in the presence of occlusion, scale change, rotation, etc. Also, it has to carefully adapt to the changes in appearances of the target persons due to the lighting variations. 

Currently, the prominent and intelligent trend in the vision-based person following is using robust visual object trackers, which are integrated into the robots to provide them with perception ability. So, the performance of a person following robot highly depends on the robustness and speed of the object tracker. Several object trackers are available in the literature that has made tremendous progress by leveraging the expressive power of the deep learning algorithms \cite{ma2015hierarchical, mayer2021learning, chen2020siamese, ma2020rpt, zhao2021trtr, yan2021learning, chen2021transformer}. Furthermore, their progress has been fueled by several popular benchmark datasets that are well established in covering diverse real-world tracking challenges like rotation, occlusion, low-resolution, small-scale objects, etc \cite{muller2018trackingnet, fan2019lasot, huang2019got, wu2013online}.

\begin{figure}[t]
    \centering
    \includegraphics[width=.95\linewidth]{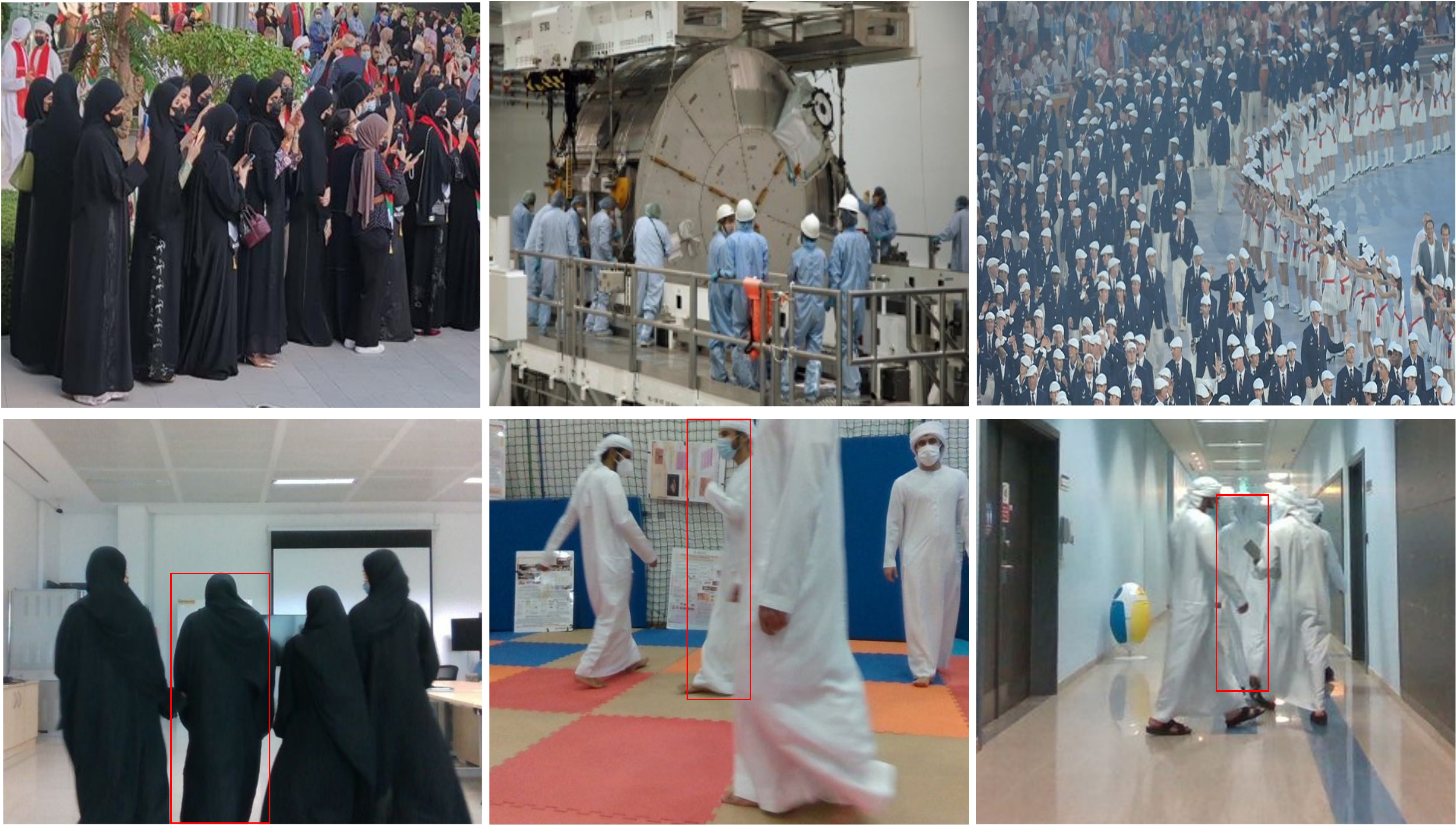}
    \caption{Sample images showing uniform crowd. $1^{st}$ row shows common crowd appearance in real-life and $2^{nd}$ row shows person following experimentation in uniform crowd environment.}
    \vspace{-7mm}
    \label{fig:sample}
\end{figure}

The currently available benchmark datasets rarely cover scenarios of person tracking in uniform crowd scenarios,i.e. tracking the specific person while people nearby him are wearing similar clothes. This is commonly seen in the places like industries, sports stadiums, and schools as shown in Fig. \ref{fig:sample}. These scenarios are also vastly found in the Gulf region and many Asiatic countries. Tracking a person in these environments based on the RGB image is a challenging task for the state-of-the-art (SOTA) visual trackers as they perceive the surroundings based on appearance features. Hence, tracking a person wearing the same garment in a uniform crowd becomes a difficult task compromising the performance of the person-tracking robot. To address this challenge,  we present a new tracking system that uses an attention-based module to learn deeper representative features from color and depth information.

We argue that in person tracking in the uniform crowd, the color information is not enough because the target and the non-targets all look the same. In addition, the problem becomes more challenging when the target is occluded by a similar-appearing distractor. Thus, to address these common problems and to come up with a robust person-following robot system, we integrate depth information into the robust tracking architecture\cite{yan2021learning}. The addition of extra depth information in the tracking pipeline allows the best  segregation of the background and the foreground irrespective of high similarity \cite{lukezic2019cdtb}. In summary, this work has three contributions:  
\begin{itemize}
    \item We propose a fully autonomous robot person following system with our new RGB-D tracker. It can track the person under the challenges like uniform crowd and occlusion with similar-appearing people. 
    \item We conducted intensive experiments to evaluate different SOTA trackers’ performance for a person following in a uniform crowd environment.
    \item We introduce a new person tracking dataset that includes the challenges of a uniform crowd.
\end{itemize}


\section{METHODOLOGY}
\label{sec:methodology}
\begin{figure}
    \centering
    \includegraphics[width=\linewidth]{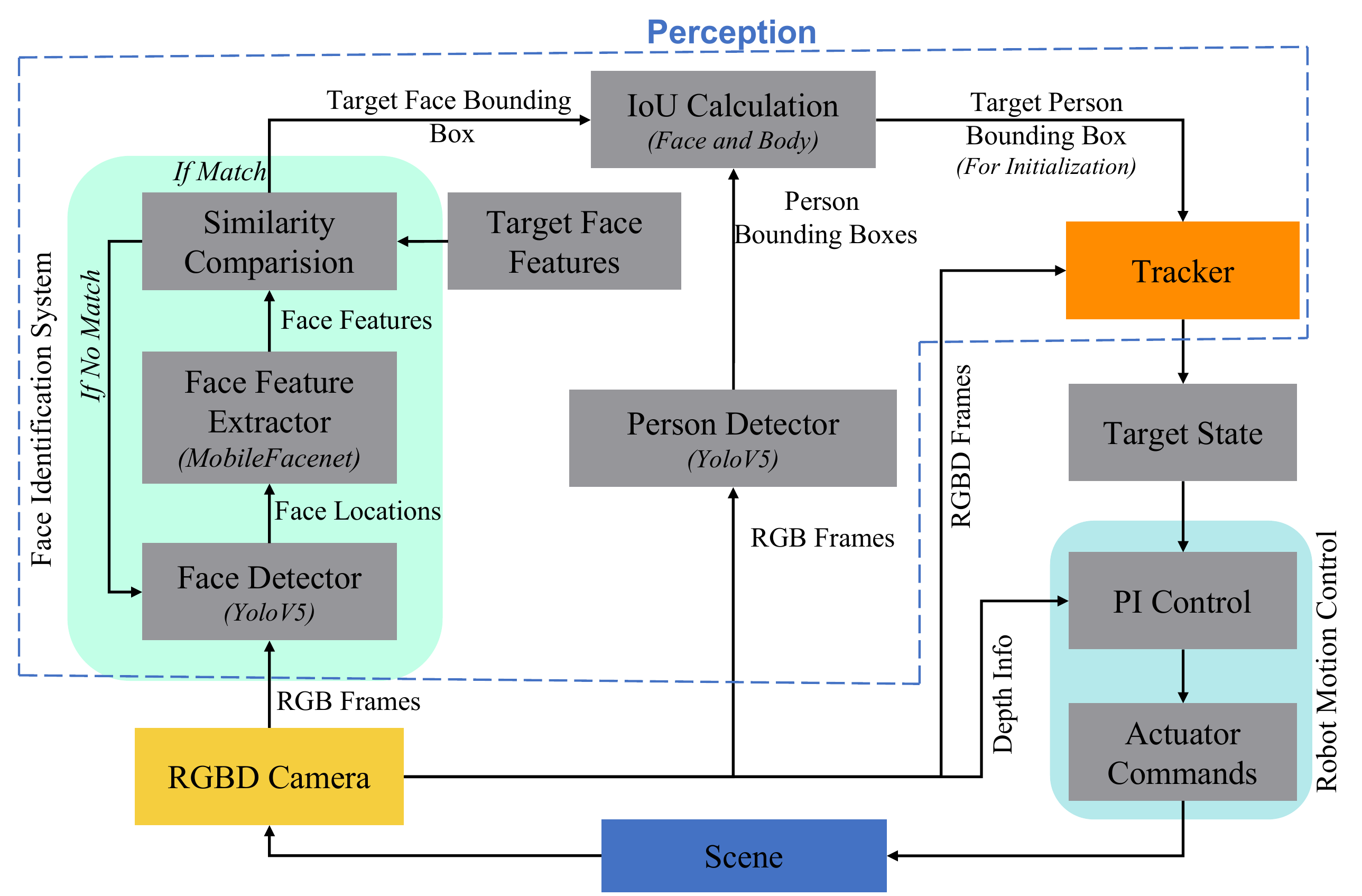}
    \caption{System Block Diagram}
    \vspace{-7mm}
    \label{fig:block_diagram}
\end{figure}

This section describes the pipeline and components of our designed robot person following system as shown in Figure. \ref{fig:block_diagram}.Our proposed robot person following system is a fully autonomous system containing two main components: the perception module (which includes face identification, person detection, and tracking) and motion control of the robot. Section \ref{subsec:perception} explains the process of how the robot continuously perceives the target person of interest, and Section \ref{subsec:control} illustrates the motion control of the robot. 

\subsection{Robot Perception}
\label{subsec:perception}
The perception module enables a robot to perceive the environment and identify its objective continuously. The main aim of the module is to allow visual servoing to work autonomously with minimal wrong behavior. Several individual components work together to allow robotic perception to work seamlessly. They are an identification module that initially identifies the specific target person's body based on face recognition and the tracking module, which tracks the target person in the crowd. 

\subsubsection{Person Identification}
\label{subsec:identification}
The faces are visible initially thus to recognize the target person's body in the scene, the overall algorithm involves three parts, face identification, person detection, and face-person matching. To achieve fast and robust person identification, we employed a pre-trained yoloV5Face \cite{qi2021yolo5face} to detect the faces and a mobile facenet \cite{chen2018mobilefacenets} for feature extraction. The mobile facenet embeds the face region into the 128-dimensional feature vector. Face verification is achieved by finding the smallest euclidean distance between the feature vectors of the known target face and detected faces with a threshold of 0.9.

Lastly, for locating the whole body of the target person in the scene, we have used a pre-trained yoloV5 detector \cite{glenn_jocher_yolov5} as a person detector to detect the people and then computed the Intersection Over Union between the regions of the detected people and the identified target face. The body with the highest IOU is considered the target person's body and is used for tracking.

\subsubsection{Tracking}
\label{subsec:tracking}
\begin{figure}
    \centering
    \includegraphics[width=0.8\linewidth]{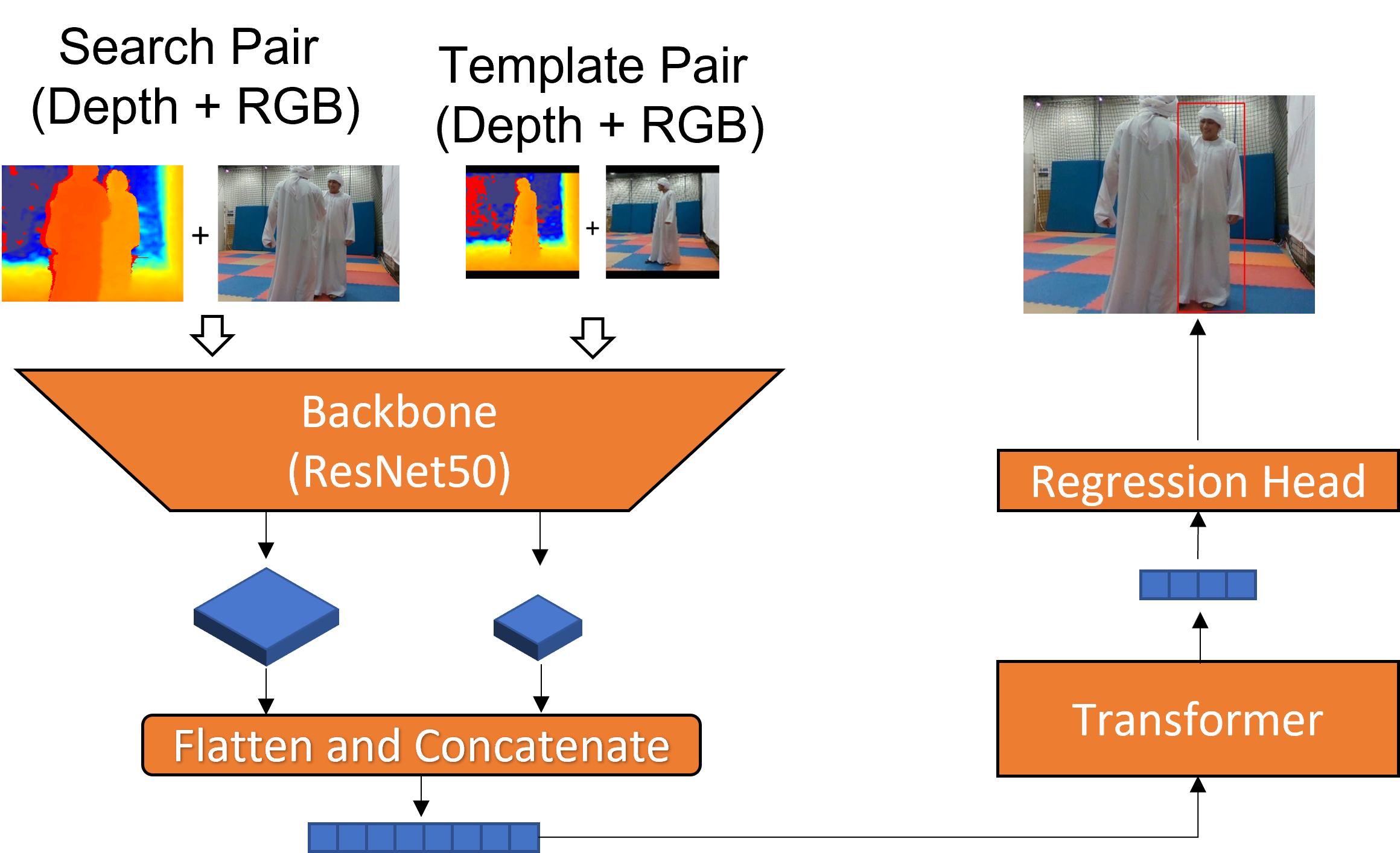}
    \caption{Proposed RGBD tracker fusing RGB and depth data}
    \vspace{-7mm}
    \label{fig:architecture}
\end{figure}

For effective person tracking in the uniform crowd environment, we utilize depth information along with RGB. We argue that RGB only is not sufficient to discriminate the visual features in a uniform crowd environment. We propose a new tracker model, which is named Deep Tracking with RGB-D (\textbf{DTRD}),  that fuses the RGB and depth images for tracking. The tracker finds discriminative features from RGB and depth together to enhance its ability to distinguish the foreground target from the highly similar background distractors. Also, the tracker architecture adds a transformer module to enhance the feature learning and give attention to a specific target of interest as shown in Fig. \ref{fig:architecture}. The model contains three main components: a convolutional backbone for feature extraction, a transformer module for attention-based feature learning, and a bounding box regression head to predict the location of the target. 

\textbf{Backbone:} The backbone adopts widely used vanilla ResNet \cite{he2016deep} convolutional layers for feature extraction for its expressivity and generality. The input of the trackers contains two pairs of images: template image $z$ and search image $s$ both containing an RGB and depth image. The pair of RGB and depth images are merged into a 4 channel image, $z \in \mathcal{R}^{H_z \times W_z \times 4}, x \in \mathcal{R}^{H_x \times W_x \times 4}$ and then passed through the backbone.  After the backbone, the input images are projected to two feature maps $f_z \in \mathcal{R}^{H_z/s \times W_z/s \times C}$ and $f_x \in \mathcal{R}^{H_x/s \times W_x/s \times C}$ where $s, C , H, W$ are the stride, number of feature channel, height and width respectively.  

\textbf{Transformer:} The transformer contains six encoder-decoder blocks and a positional embedding layer. Before being fed into the transformer module, the feature maps are flattened and concatenated into a feature vector with a length of $H_x/s \times W_x/s + H_z/s \times W_z/s $ and a dimension of C. The transformer embeds the feature vector into a one-dimensional vector with a length of $d$.

\textbf{Regression Head:} The regression head is three fully connected layers to predict the four coordinates of the bounding box. The regression head directly predicts the top-left corner and the bottom-right corner of the target bounding box.

\subsection{Motion Control}
\label{subsec:control}
The designed robot-person following system is a visual servoing system in which the object tracker plays the most important role in locating the target. Thus, to keep the robot following the target person in an adaptive manner we have utilized a simple PI controller as the robot's motion controller. Also, we argue that a good object tracker should be able to compensate for the controller to improve the robot person following. The PI controller is defined in equation \ref{eqs:picontrol}.
\vspace{-4mm}
\begin{equation}
    u(t) = K_p e(t) + K_i \int e(t)dt
    \vspace{-2mm}
    \label{eqs:picontrol}
\end{equation}

Where the $K_p, K_i$ are the control gains, $e(t)$ is the error between the current measure and required measure, and $u(t)$ is the velocity output. For both the linear velocity and angular velocity, two PI controllers have been used.

\section{EXPERIMENT}
\label{sec:experiment}
In this section, we detail the developed dataset, tracker training, and intensive experiments performed to compare and evaluate the robot person following systems in a uniform crowd environment in real-time. Section \ref{subsec:dataset} presents the developed dataset and our tracker training detail. Section \ref{subsec:realtime} explains the real-time robot person tracking experimentation setup. Section \ref{subsec:result} and  \ref{subsec:discussion} wraps up the results of the experiment with an insightful observation.

\subsection{Dataset and Training Detail}
\label{subsec:dataset}
The primary problem for our proposed tracker was the lack of a uniform crowd tracking dataset(a similar-appearing distractor keeps on distracting in the view) that also contains depth data. Therefore, 45 uniform crowd videos (avg. 1891 frames/video) with synchronized RGB and depth data were collected and properly annotated initially with CVAT\footnote{\href{https://cvat.org/}{\textit{https://cvat.org/}}}(shown in $2^{nd}$ row of Fig.\ref{fig:sample}). The dataset records several scenarios such as occlusion, illumination variations, motion blur, etc.   

We trained our tracker with $70\%$ of the dataset with initial weights from the STARK \cite{yan2021learning} spatial model. To properly embed depth, the first convolution layer of the backbone was modified to take four-channel image input and was trained from scratch. The training was done for 10 epochs with IOU and L1 loss using an ADAMW optimizer with a learning rate of 0.0001 for the model and 0.00001 for the backbone. The trained tracker was evaluated on the remaining $30\%$ of the dataset to confirm tracking ability. 

\subsection{Real-time Tracking}
\label{subsec:realtime}
To compare the system following the target, we selected 2D SOTA trackers(DiMP\cite{bhat2019learning}, ATOM\cite{danelljan2019atom}, SiamBAN\cite{chen2020siamese}, STARK\cite{yan2021learning}, KeepTrack\cite{mayer2021learning}) that only use color image and our proposed 3D tracker that uses color and depth image. 

\subsubsection{Experiment Setup}
\begin{figure*}[t]
    \centering
    \begin{subfigure}{.245\textwidth}
    \centering
    \includegraphics[width=\linewidth]{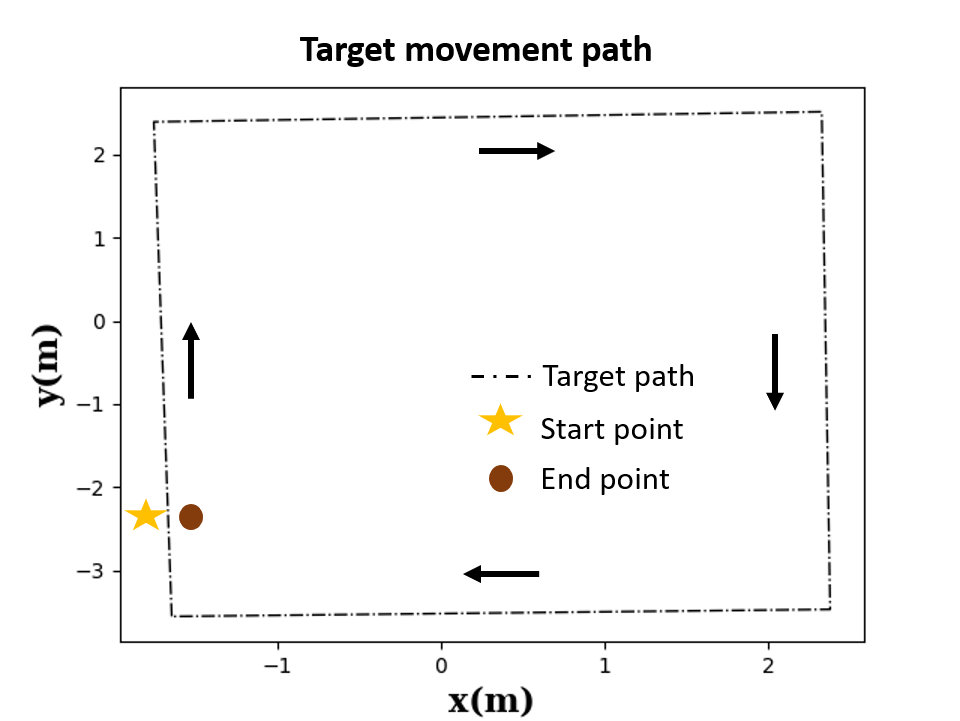}
    \caption{Target moving path}
    \label{fig:target_moving_path}
    \end{subfigure}
    \begin{subfigure}{.245\textwidth}
    \centering
    \includegraphics[width=\linewidth]{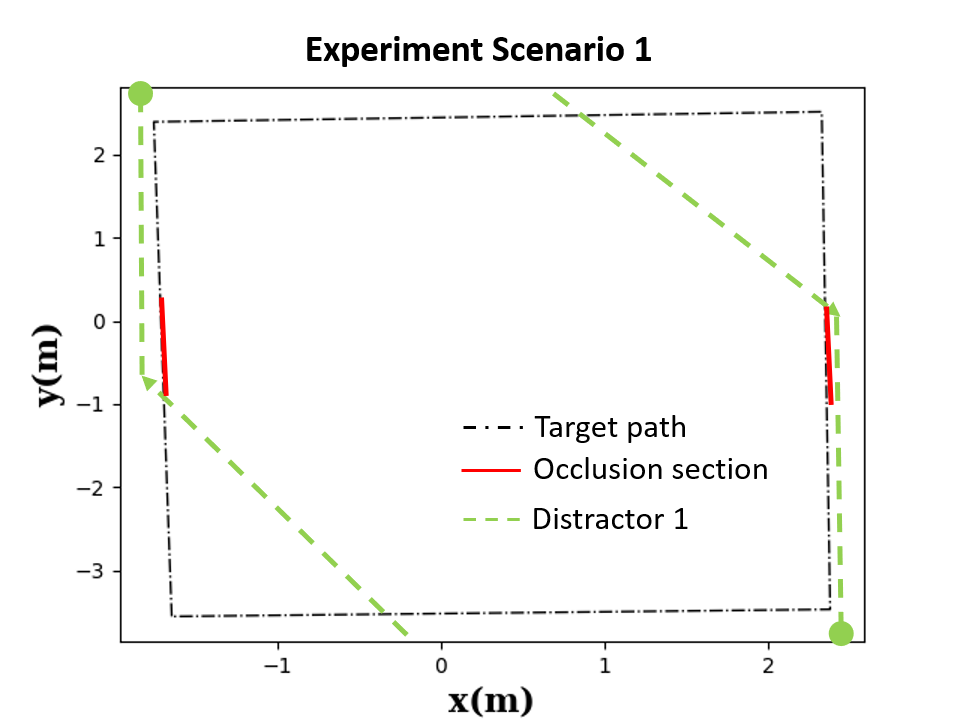}
    \caption{Single distractor}
    \label{fig:single_distractor_path}
    \end{subfigure}
    \begin{subfigure}{.245\textwidth}
    \centering
    \includegraphics[width=\linewidth]{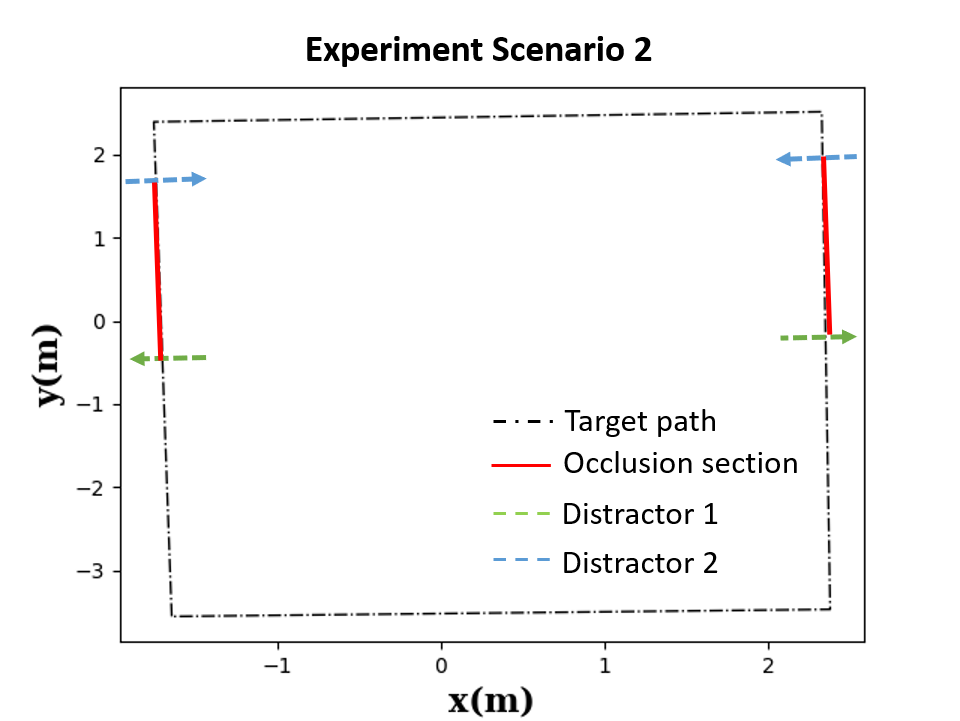}
    \caption{Two distractors (cross walking)}
    \label{fig:two_distractor_cross_path}
    \end{subfigure}
    \begin{subfigure}{.245\textwidth}
    \centering
    \includegraphics[width=\linewidth]{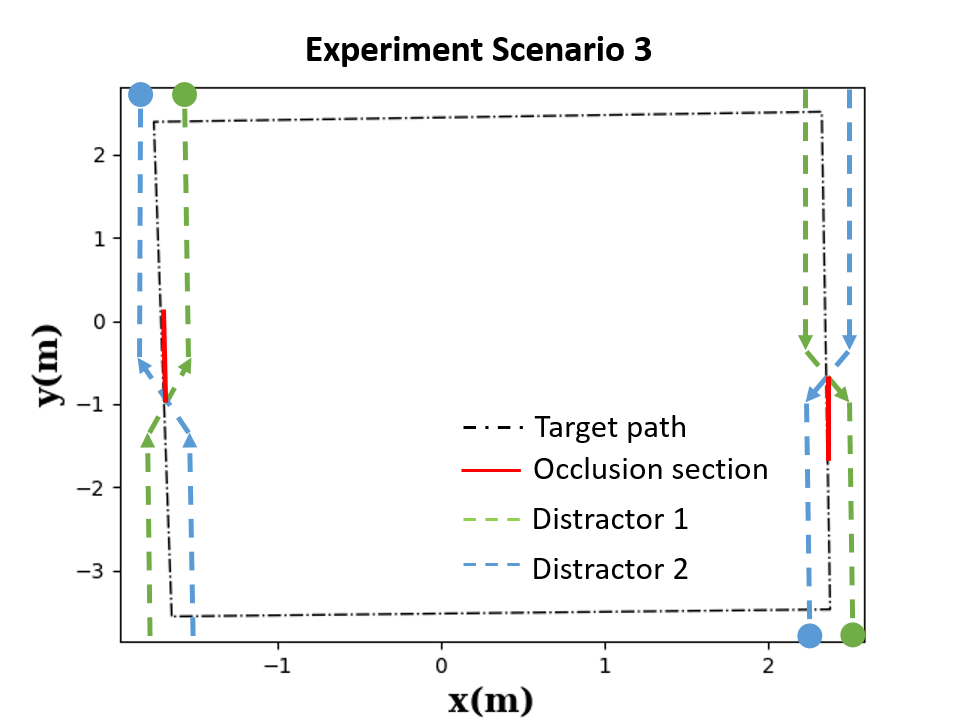}
    \caption{Two distractors (parallel walking)}
    \label{fig:two_distractor_parallel_path}
    \end{subfigure}
    \caption{Pre-defined path for experimentation (Top view). The red solid line is where the target is occluded by the distractor/s.}
    \vspace{-3mm}
    \label{fig:path}
    
\end{figure*}

We used the pioneer3 AT robot as the mobile platform with ZEDm RGB-D camera placed one meter above the robot and a laptop with GPU as the control unit. We used an OptiTrack system\footnote{\href{https://optitrack.com/}{\textit{https://optitrack.com/}}} to measure the actual positions of the robot and people in the scene.

To make an effective and fair comparison between different trackers, we set a predefined rectangular path of the target person's movement as shown in Fig. \ref{fig:target_moving_path} and a trial is complete once the person reaches its initial position. The initial positions of the robot and target person are fixed at the left bottom corner (of the Fig. \ref{fig:target_moving_path}) with the robot placed at the corner and the person is 2 meters in front of it.

For studying the impact of different levels of the uniform crowd on a person following system,  we conducted experiments by adding a different number of distractors in the scene to examine the system performance with different trackers. Designed experiments include no distractor scene which provides a baseline evaluation of the system; One distractor scene where distractor walk cross to occlude the target as shown in Fig. \ref{fig:single_distractor_path}; Two distractor scenes where distractors walk cross or parallel to intervene the robot as shown in Fig. \ref{fig:two_distractor_cross_path} and \ref{fig:two_distractor_parallel_path}.



To avoid discrepancies during metric evaluation the distractor stops once the robot starts following him while the target person continues its path. A complete experiment protocol has been conducted as in Algorithm \ref{alg:realtime_experiment} to robustly and fairly examine the system's ability to follow the target.
\vspace{-2mm}
\begin{algorithm}
\caption{: Real-time experiment protocol} 
\begin{algorithmic}[1]
\For{subject = A, B}
    \For{tracker = DiMP, ATOM, SiamBAN, STARK, KeepTrack, DTRD} 
        \For{distractor = 0, 1, 2} 
            \For{trial = 1, 2, 3}
                \State Perform robot person tracking experiment
            \EndFor
        \EndFor
    \EndFor
\EndFor
\end{algorithmic}
\label{alg:realtime_experiment}
\end{algorithm}
\vspace{-3mm}

\subsubsection{Experiment Evaluation Metric} 
\label{subsec:experiment_evaluation_metric}
In previous research, we found the majority of work used distances error and trajectory matching to quantitatively and qualitatively evaluate person following. These metrics are insufficient to evaluate the systems' performance, especially in the case of the intervention from the distractors. Hence, we also add a new metric, following success, in the experiment for robust and quantitative evaluation as listed in Table \ref{tab:metrics}.

\begin{table*}[!t]
\centering
\caption{Evaluation metrics for robot person following}
\label{tab:metrics}
\begin{tabular}{p{0.25\textwidth}|p{0.65\textwidth}}
\hline
\multicolumn{1}{c|}{Metrics} & \multicolumn{1}{c}{Descriptions}                                                                                                                                           \\ \hline
Distance Error (DE)          & The difference in measurement from the tracker and the ground truth from the OptiTrack. Higher is the worst.                                  \\ \hline
Following Success (FS)       & Duration of the robot is successfully following the target person.  FS = {[}0, 1{)}. Higher is the best.                                                 \\ \hline
Frame rate Per Second (FPS)  & Frame processing time for the tracker which indicates the system's response time.
\\ \hline
\end{tabular}
\vspace{-3mm}
\end{table*}

\subsection{Results}
\label{subsec:result}
We report results on four evaluation metrics based on intensive experimentation.

\begin{figure*}[t]
    \centering
    \begin{subfigure}{.3\textwidth}
    \centering
    \includegraphics[width=\linewidth]{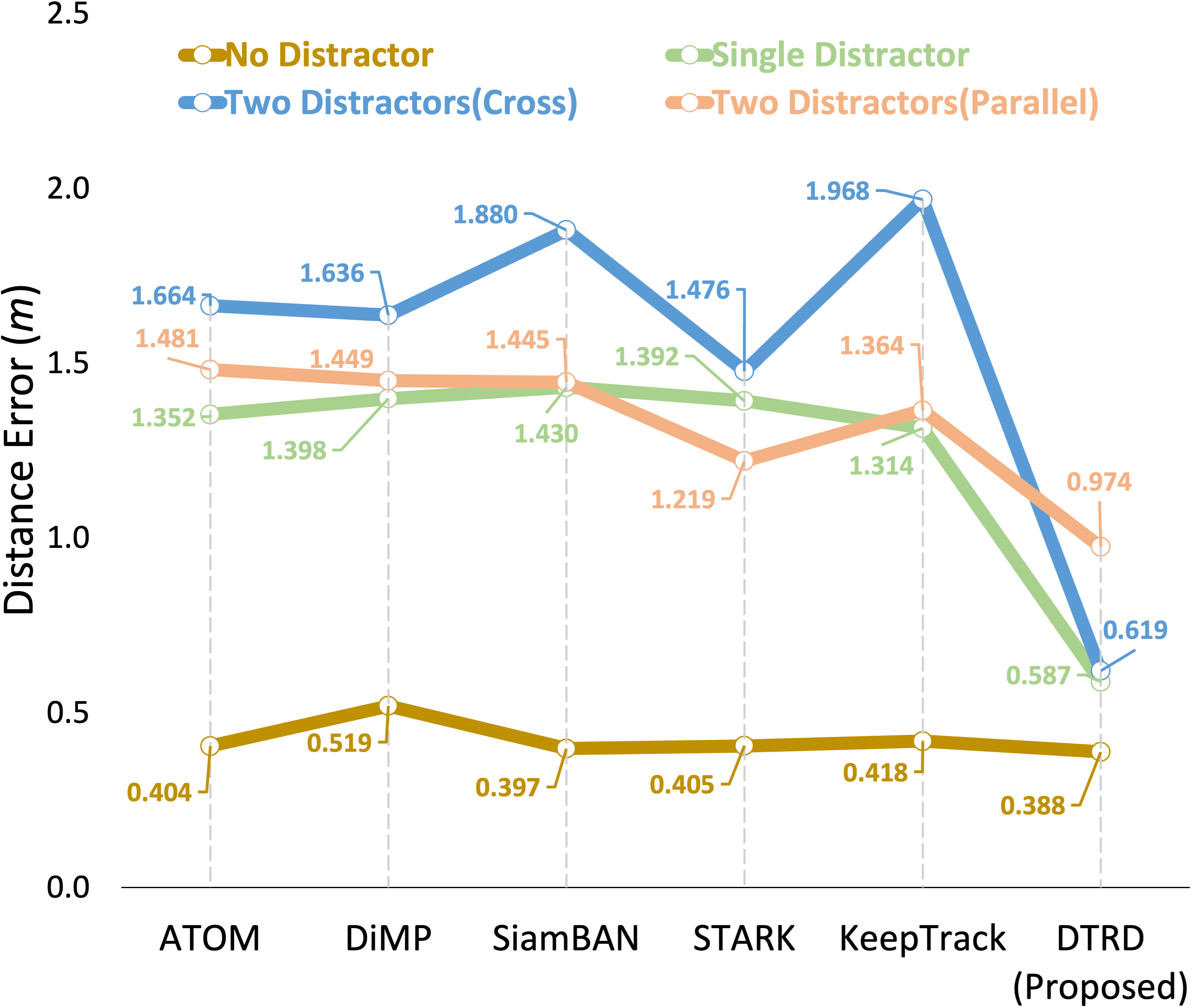}
    \caption{Distance error}
    \label{fig:distanceerror_curve}
    \end{subfigure}
    \begin{subfigure}{.3\textwidth}
    \centering
    \includegraphics[width=\linewidth]{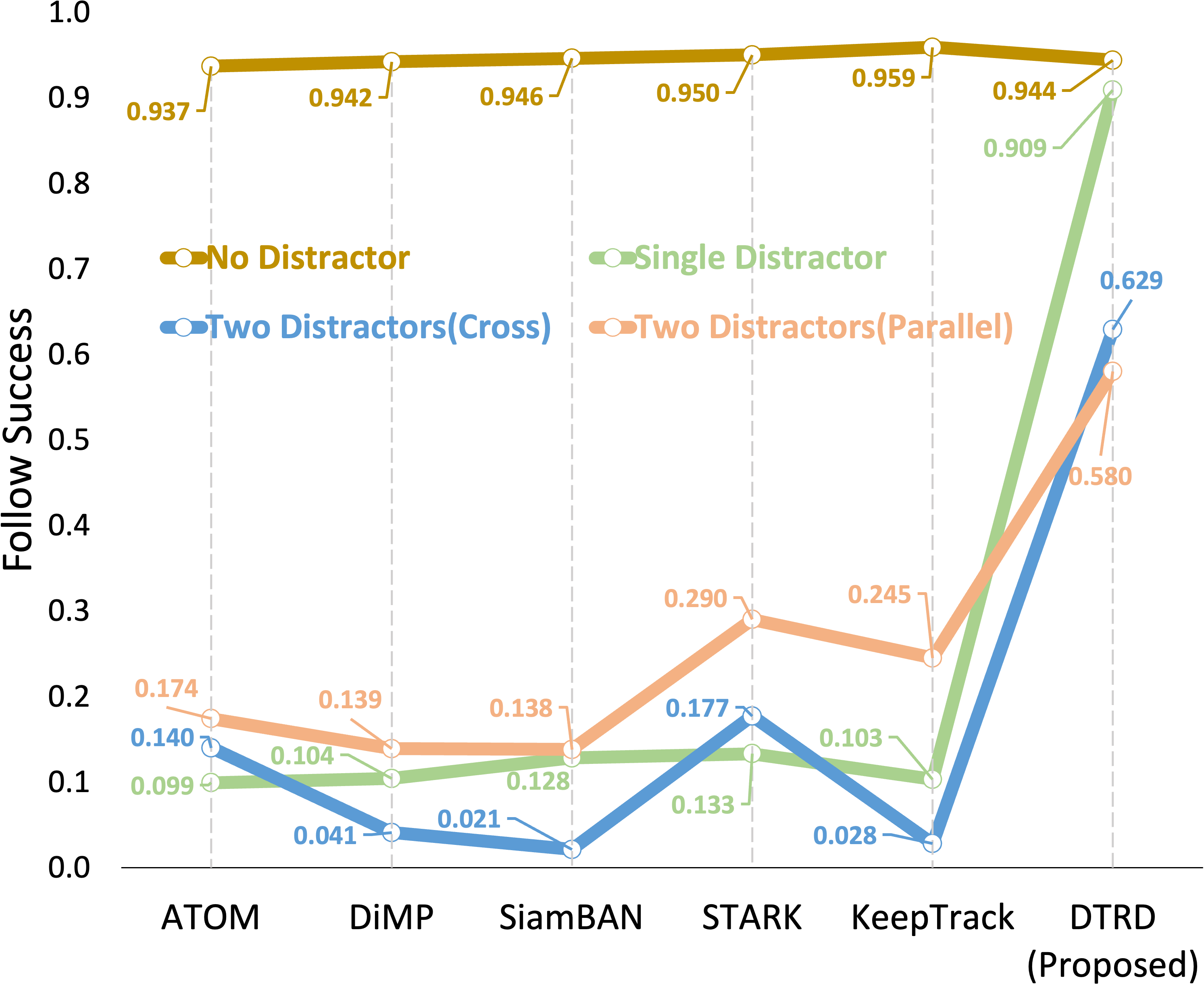}
    \caption{Person Following Success}
    \label{fig:fs_curve}
    \end{subfigure}
    \begin{subfigure}{.3\textwidth}
    \centering
    \includegraphics[width=\linewidth]{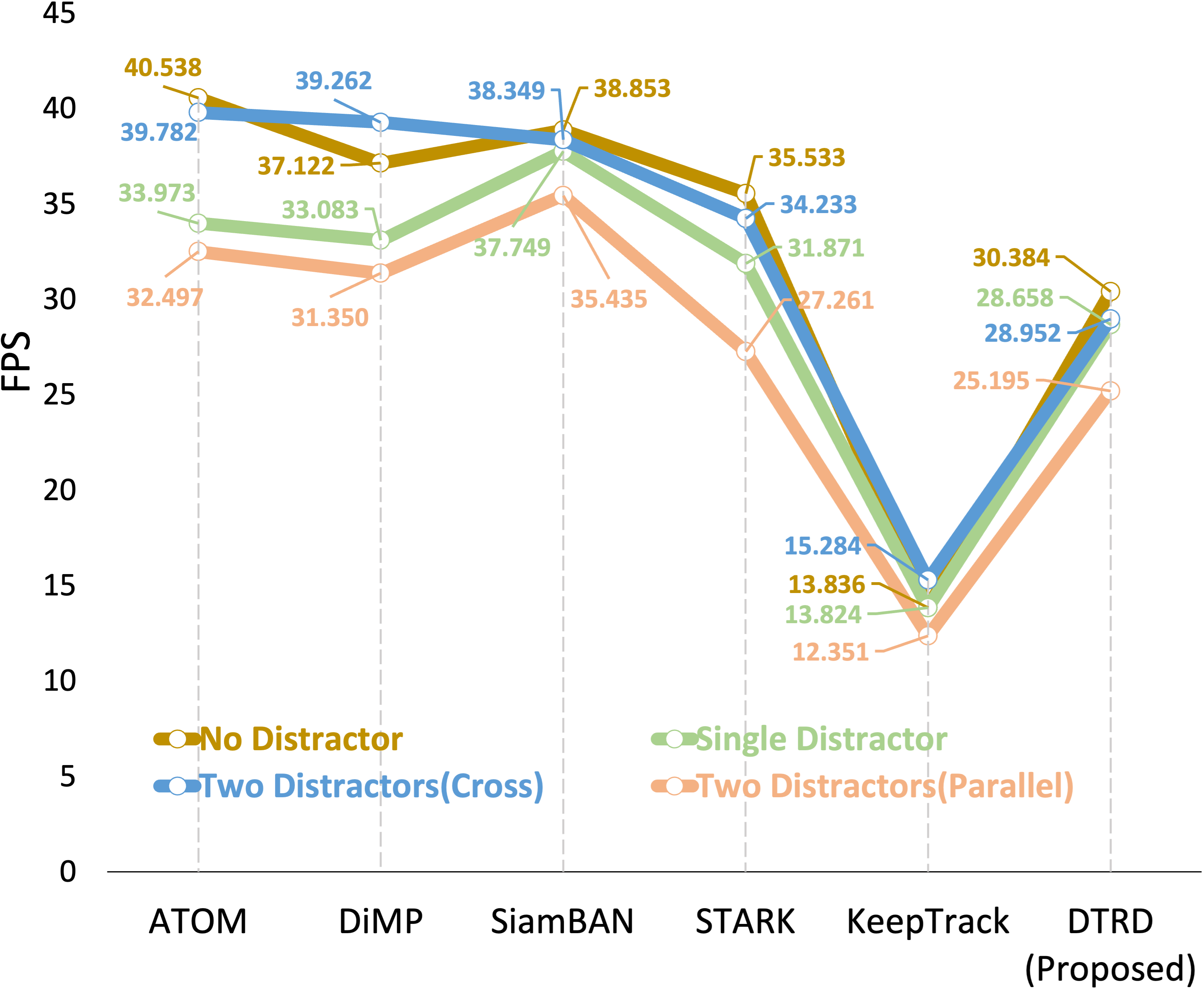}
    \caption{Frame Processing Time}
    \label{fig:fps_curve}
    \end{subfigure}
    \caption{Experiment results}
    \vspace{-5mm}
\end{figure*}

\noindent\textbf{\textit{DE} Result:}
$DE$ quantifies how well the tracker predicts the target's state and signifies the error in distance measured from the tracker to the actual distance measured from the OptiTrack. From Fig. \ref{fig:distanceerror_curve} we can see that the overall $DE$ is around $0.5 m$ when there is no distractor in the field. But for all the SOTA trackers the errors have increased significantly with distractor presented scenarios making an average distance error greater than $1.2 m$. However, in our proposed DTRD tracker, it is around $0.6 m$ only for the two-distractor(cross) scenario. Whereas for two-distractor(parallel) it is slightly higher than other scenarios but still has the lowest error compared to all other trackers.   


\noindent\textbf{\textit{FS} Result:}
$FS$ evaluates how well the tracker can drive the robot following the target person. This metric is calculated by comparing the distances traveled by the robot and the target person. The metric is derived based on the fact that, if the robot losses the target early, the distance traveled by the robot is short resulting in a lower $FS$ value. However, a higher $FS$ value signifies that the tracker is providing a robust perception to the robot despite any distractions. In Fig. \ref{fig:fs_curve}, when there are no distractors the $FS$ value for all the trackers is very high. However, when there are distractors all the 2D trackers have an $FS$ value of around 0.2 indicating they lost the target very early. But our proposed tracker has a higher $FS$ value than other trackers, showcasing robustness to the distractor.


\noindent\textbf{\textit{FPS} Result:}
This metric reveals the frame processing time of trackers which has a great impact on the response time and stability of the robot in real-time ($FPS$ should be above $20$). From Fig. \ref{fig:fps_curve}, we can see that all of the trackers fulfilled the constrained, except for KeepTrack (avg $FPS$ $13$).


\subsection{Discussion} 
\label{subsec:discussion}
From the experimental results, it is evident that the visual-based robot person following system is able to keep track of the target person, referring to the reported higher performance of all trackers in the no-distractor scenario. However, the system performance significantly degrades when the distractors intervene. But our proposed tracker provides better perception ability than other 2D trackers. The reasons are two folds. Firstly, all the SOTA 2D trackers are only relying on appearance thus failing in uniform crowd scenarios and our DTRD tracker improved performance by exploiting the depth as well as using an attention-based module to learn better features from the images. Secondly, we observed that our DTRD tracker performs well in the uniform crowd when there is some distance between the target and distractors, but can get distracted otherwise. This is evidenced by the $FS$ value (around 0.6) of the DTRD in two-distractor scenarios showing the DTRD resisted the distractor's first intervention but failed at the second one because the gap was small.

Nevertheless, a more advanced design of trackers is required in the future to overcome the challenges of a uniform crowd environment. Also, trajectory prediction using visual cues is needed to make the robot handle the target out-of-view situation. It is to note that there is a great potential for depth to solve the challenges of tracking.

\section{Conclusion}
\label{sec:conclusion}
In this work, we present a new person tracking method exploiting the RGB and depth data that can overcome the deceptive challenge introduced by similar-appearing distractors. The proposed system is robust to different challenges of occlusion and distraction by the uniform crowd and thus provides resilient perception ability to the robot. Our tracking method also outperforms the SOTA 2D trackers in three different robust quantitative evaluation metrics.





\bibliographystyle{IEEEtran}
\bibliography{IEEEabrv,Bibliography}

\begin{thebibliography}{10}
\providecommand{\url}[1]{#1}
\csname url@rmstyle\endcsname
\providecommand{\newblock}{\relax}
\providecommand{\bibinfo}[2]{#2}
\providecommand\BIBentrySTDinterwordspacing{\spaceskip=0pt\relax}
\providecommand\BIBentryALTinterwordstretchfactor{4}
\providecommand\BIBentryALTinterwordspacing{\spaceskip=\fontdimen2\font plus
\BIBentryALTinterwordstretchfactor\fontdimen3\font minus
  \fontdimen4\font\relax}
\providecommand\BIBforeignlanguage[2]{{%
\expandafter\ifx\csname l@#1\endcsname\relax
\typeout{** WARNING: IEEEtran.bst: No hyphenation pattern has been}%
\typeout{** loaded for the language `#1'. Using the pattern for}%
\typeout{** the default language instead.}%
\else
\language=\csname l@#1\endcsname
\fi
#2}}

\bibitem{islam2019person}
M.~J. Islam, J.~Hong, and J.~Sattar, ``Person-following by autonomous robots: A
  categorical overview,'' \emph{IJRR}, vol.~38, no.~14, pp. 1581--1618, 2019.

\bibitem{VOT_TPAMI}
M.~Kristan, J.~Matas, A.~Leonardis, T.~Vojir, R.~Pflugfelder, G.~Fernandez,
  G.~Nebehay, F.~Porikli, and L.~\v{C}ehovin, ``A novel performance evaluation
  methodology for single-target trackers,'' \emph{IEEE PAMI}, vol.~38, no.~11,
  pp. 2137--2155, Nov 2016.

\bibitem{kristan2018sixth}
M.~Kristan, A.~Leonardis, J.~Matas, M.~Felsberg, R.~Pflugfelder,
  L.~ˇCehovin~Zajc, T.~Vojir, G.~Bhat, A.~Lukezic, A.~Eldesokey,
  \emph{et~al.}, ``The sixth visual object tracking vot2018 challenge
  results,'' in \emph{ECCV}, 2018.

\bibitem{kristan2019seventh}
M.~Kristan, J.~Matas, A.~Leonardis, M.~Felsberg, R.~Pflugfelder, J.-K.
  Kamarainen, L.~ˇCehovin~Zajc, O.~Drbohlav, A.~Lukezic, A.~Berg,
  \emph{et~al.}, ``The seventh visual object tracking vot2019 challenge
  results,'' in \emph{IEEE ICCV Workshops}, 2019.

\bibitem{kristan2020eighth}
M.~Kristan, A.~Leonardis, J.~Matas, M.~Felsberg, R.~Pflugfelder, J.-K.
  K{\"a}m{\"a}r{\"a}inen, M.~Danelljan, L.~{\v{C}}. Zajc,
  A.~Luke{\v{z}}i{\v{c}}, O.~Drbohlav, \emph{et~al.}, ``The eighth visual
  object tracking vot2020 challenge results,'' in \emph{ECCV}.\hskip 1em plus
  0.5em minus 0.4em\relax Springer, 2020.

\bibitem{Kristan2021ninth}
M.~Kristan, J.~Matas, A.~Leonardis, M.~Felsberg, R.~Pflugfelder, J.-K.
  Kamarainen, H.~J. Chang, M.~Danelljan, L.~\v{C}ehovin Zajc,
  A.~Luke\v{z}i\v{c}, O.~Drbohlav, J.~Kapyla, G.~Hager, S.~Yan, J.~Yang,
  Z.~Zhang, G.~Fernandez, and et. al., ``The ninth visual object tracking
  vot2021 challenge results,'' 2021.

\bibitem{chen2017integrating}
B.~X. Chen, R.~Sahdev, and J.~K. Tsotsos, ``Integrating stereo vision with a
  cnn tracker for a person-following robot,'' in \emph{IEEE ICCV}, 2017.

\bibitem{chen2017person}
B.~X. Chen, R.~Sahdev, and J.~Tsotsos, ``Person following robot using selected
  online ada-boosting with stereo camera,'' in \emph{IEEE CRV}, 2017.

\bibitem{ma2015hierarchical}
C.~Ma, J.-B. Huang, X.~Yang, and M.-H. Yang, ``Hierarchical convolutional
  features for visual tracking,'' in \emph{IEEE ICCV}, 2015.

\bibitem{mayer2021learning}
C.~Mayer, M.~Danelljan, D.~P. Paudel, and L.~Van~Gool, ``Learning target
  candidate association to keep track of what not to track,'' \emph{arXiv
  preprint arXiv:2103.16556}, 2021.

\bibitem{chen2020siamese}
Z.~Chen, B.~Zhong, G.~Li, S.~Zhang, and R.~Ji, ``Siamese box adaptive network
  for visual tracking,'' in \emph{IEEE CVPR}, 2020.

\bibitem{ma2020rpt}
Z.~Ma, L.~Wang, H.~Zhang, W.~Lu, and J.~Yin, ``Rpt: Learning point set
  representation for siamese visual tracking,'' in \emph{ECCV}.\hskip 1em plus
  0.5em minus 0.4em\relax Springer, 2020.

\bibitem{zhao2021trtr}
M.~Zhao, K.~Okada, and M.~Inaba, ``Trtr: Visual tracking with transformer,''
  \emph{arXiv preprint arXiv:2105.03817}, 2021.

\bibitem{yan2021learning}
B.~Yan, H.~Peng, J.~Fu, D.~Wang, and H.~Lu, ``Learning spatio-temporal
  transformer for visual tracking,'' \emph{arXiv preprint arXiv:2103.17154},
  2021.

\bibitem{chen2021transformer}
X.~Chen, B.~Yan, J.~Zhu, D.~Wang, X.~Yang, and H.~Lu, ``Transformer tracking,''
  in \emph{IEEE CVPR}, 2021.

\bibitem{muller2018trackingnet}
M.~Muller, A.~Bibi, S.~Giancola, S.~Alsubaihi, and B.~Ghanem, ``Trackingnet: A
  large-scale dataset and benchmark for object tracking in the wild,'' in
  \emph{ECCV}, 2018.

\bibitem{fan2019lasot}
H.~Fan, L.~Lin, F.~Yang, P.~Chu, G.~Deng, S.~Yu, H.~Bai, Y.~Xu, C.~Liao, and
  H.~Ling, ``Lasot: A high-quality benchmark for large-scale single object
  tracking,'' in \emph{IEEE CVPR}, 2019.

\bibitem{huang2019got}
L.~Huang, X.~Zhao, and K.~Huang, ``Got-10k: A large high-diversity benchmark
  for generic object tracking in the wild,'' \emph{IEEE PAMI}, 2019.

\bibitem{wu2013online}
Y.~Wu, J.~Lim, and M.-H. Yang, ``Online object tracking: A benchmark,'' in
  \emph{IEEE CVPR}, 2013.

\bibitem{lukezic2019cdtb}
A.~Lukezic, U.~Kart, J.~Kapyla, A.~Durmush, J.-K. Kamarainen, J.~Matas, and
  M.~Kristan, ``Cdtb: A color and depth visual object tracking dataset and
  benchmark,'' in \emph{Proc. IEEE ICCV}, 2019.

\bibitem{qi2021yolo5face}
D.~Qi, W.~Tan, Q.~Yao, and J.~Liu, ``Yolo5face: Why reinventing a face
  detector,'' \emph{arXiv preprint arXiv:2105.12931}, 2021.

\bibitem{chen2018mobilefacenets}
S.~Chen, Y.~Liu, X.~Gao, and Z.~Han, ``Mobilefacenets: Efficient cnns for
  accurate real-time face verification on mobile devices,'' in \emph{CCBR},
  2018.

\bibitem{glenn_jocher_yolov5}
\BIBentryALTinterwordspacing
G.~J. et~al., ``ultralytics/yolov5,'' Oct. 2020. [Online]. Available:
  \url{https://doi.org/10.5281/zenodo.4154370}
\BIBentrySTDinterwordspacing

\bibitem{he2016deep}
K.~He, X.~Zhang, S.~Ren, and J.~Sun, ``Deep residual learning for image
  recognition,'' in \emph{IEEE CVPR}, 2016.

\bibitem{bhat2019learning}
G.~Bhat, M.~Danelljan, L.~V. Gool, and R.~Timofte, ``Learning discriminative
  model prediction for tracking,'' in \emph{IEEE ICCV}, 2019.

\bibitem{danelljan2019atom}
M.~Danelljan, G.~Bhat, F.~S. Khan, and M.~Felsberg, ``Atom: Accurate tracking
  by overlap maximization,'' in \emph{IEEE CVPR}, 2019.

\end{thebibliography}

\end{document}